# HYBRID FEATURE BASED SLAM PROTOTYPE

V.I.Mebin Jose[1], D.J Binoj[2]
[1]Student -MS university ,Tirunelveli,Tamil Nadu,India
[2]Project co-ordinator - jitec technologies ,Karungal,Tamil Nadu,India

**ABSTRACT**

The development of data innovation as of late and the expanded limit, has permitted the acquaintance of artificial vision connected with SLAM, offering ascend to what is known as Visual SLAM. The objective of this paper is to build up a route framework dependent on Visual SLAM to get a robot to a fundamental and new condition, have the capacity to set and make a three-dimensional guide thereof, utilizing just as sources of info recording your way with a stereo vision camera. The consequence of this analysis is that the framework Visual SLAM together with the combination of Fast SLAM (combination of kalman with particulate filter and SIFT) perceive and recognize characteristic points in images so adequately exact and unambiguous. This framework uses MATLAB, since its adaptability and comfort for performing a wide range of tests. The program has been tested by inserting a prerecorded video input with a camera stereo in which a course is done by an office environment. The algorithm initially locates points of interest in a stereo frame captured by the camera. These will be located in 3D and they associate an identification descriptor. In the next frame, the camera likewise identified points of interest and it will be compared which of them have been previously detected by comparing their descriptors. This process is known as "data association" and its successful completion is fundamental to the SLAM algorithm. The position data of the robot and points interest stored in data structures known as "particles" that evolve independently. Its management is very important for the proper functioning of the algorithm Fast SLAM. The results are found to be satisfactory.

**Introduction to Visual SLAM**

Simultaneous Localization and Mapping (SLAM), [1][2] in unknown environments is a hot issue of mobile robot research field [3][4]. SLAM of traditional mobile robot is obtained by sonar sensor and laser radar, etc. distance sensors. Because of the low resolution of these distance sensors, in a complex environment, [5] [6] it is difficult to achieve the desired results owing to the high uncertain observational data . Several papers consider the feature-based SLAM problem using camera images [7] [8].

Visual l SLAM (Simulation Modeling for Alternative Language) is a language graphic oriented simulation systems and processes. SLAM was developed in 1979 by Dennis and Alan Pedge Pritsker and was distributed by Pritsker Corporation. SLAM part of which it is oriented to the processes used a lattice structure [9] composed of nodes and branches symbols such as queues, servers and decision points. Modeling means incorporating these symbols to a network model that represents the system and where various entities pass through the network. SLAM contains a processor that converts the visual representation of the system to a set of instructions. The Visual SLAM II [10] requires a modeling program to build a network representing the system / process (problem), to make a compilation and finally executed. Once implementation has been completed successfully, SLAM II produces some forms of output. The SLAM used for for endoscopic capsule robot operations by monocular visual odometry (VO) method [11] used for the application of the deep recurrent convolutional neural networks (RCNNs) for the visual odometry task, covariance [12] based simultaneous detection also there is the SLAM In the last decade, many medical companies and research groups have tried to convert passive capsule endoscopes as an emerging and minimally invasive diagnostic technology into non invasive technology [13]

There are three basic components of SLAM; these components are entities, nodes and arcs. Entities are the units that will be transformed over time. The entities are not part of the network but they move through it. All entities flowing through the network need to be identical. Attributes To determine the specific function of an entity and distinguish it from other, and thus allow the modeler data logging, SLAM II provides each entity a set of attributes (ATRIB's). The ATRIB's are variable storage locations ranging attached to each entity and in which the modeler can store and recall data.

**Visual SLAM types**

LOS algorithms SLAM (Simultaneous Location and Mapping) are the future of robotics. The ability to localize in an environment and mapping is essential for creating truly autonomous robots, which are not only intended for activities specific, if they are able to adapt to circumstances and to what surrounds him. Get solve this problem will take off service robotics, where robots leave their traditional industrial environment and begin to be used for direct support to humans in all tasks directly affecting their quality of life improve. Currently This concept is being applied in domestic robots of all types (such as cleaner Figure 1), surveillance robots , robots for space exploration, etc.

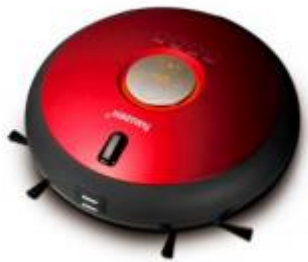

**Figure 1: Robot cleaner Samsung Hauzen**

A fundamental aspect of a system based on SLAM is used to observe the environment, since the abilities and functions depend largely robot that election. The advance of information technology in recent years and the increased capacity, has allowed the introduction of artificial vision applied to SLAM, giving rise to what is known as Visual SLAM. By employing cameras Global information that surrounds the robot is obtained, not only in one dimension as it happens with the usual radar systems, sonar or ultrasound, but has the disadvantage that the information is difficult to process and analyze. In this project a view camera is used stereo, imitating the human eye, which has two separate chambers allow "see" in three dimensions.

**The Problem of SLAM**

The problem of SLAM (in Spanish: Location and armed with simultaneous maps) applies when the robot does not have access to a map of the environment nor know their position in the same. In Figure 1.2, for example, a robot used for rescue is observed in an area where the human cannot access because of the high levels of radioactivity, in this case the robot does not have knowledge of its location or environment. Then, the robot only has the information provided by the measurements obtained sensors and the notion of proper motion. The agent will try to obtain a map environment and simultaneously located on this map. In the context of SLAM there are several sources of uncertainty, i.e. factors that increase the difficulty of estimate the location and map the correct environment. Some of these factors are:

**Noise sensors:** The sensors used in a robot usually present noise in the data provided.

**Imprecise robot displacement:** The result of a movement of the robot is non-deterministic nature. This is because, for example, the wheels of a robot. They can slip on the ground. As a result, it is not possible to know for sure as was the actual movement of the robot as a result of a move order.

**Symmetries in the environment:** The environment on which the robot operates may have symmetries that they hinder the determination of the current position of the robot in the map prepared.

**Partial Observability:** The absence of a mechanism for global vision of the environment difficult to estimate the position and the construction of map.

**Dynamic environment:** If you work in a dynamic environment, changes in the environment will hinder the process of estimating the position because the map prepared it may be outdated. After adapting to this uncertainty, much of SLAM solutions raise the solution estimating the position of the robot and the environment map as distributions probability.

After adapting to this uncertainty, much of SLAM solutions raise the solution estimating the position of the robot and the environment map as distributions probability.

Thus the aim of this project is to develop a navigation system based on Visual SLAM to get a robot to a basic and unfamiliar environment, be able to placed and create a three-dimensional map thereof, using only as inputs recording your path with a stereo vision camera is coupled to said robot. To achieve this goal was made first a thorough study of the state Art SLAM algorithms and detection algorithms and description of points of interest. The result of this analysis is that the system SLAM Visual most complete to date based on the combination of Fast SLAM, SLAM algorithm that combines traditional filter Kalman with particulate filters, taking advantage of both, and SIFTS, an algorithm machine vision able to recognize and identify characteristic points in images so sufficiently precise and unambiguous. Once decided the system has proceeded to its implementation in Matlab, since its flexibility and convenience when scheduling makes it ideal for all kinds of tests. Nowdays Visual-Magnetic Sensor Fusion Approach also helpful in the field of visual slam and medical [15] [16]

**Conclusion**

This project helps to improve the SLAM performance by addition of features to predict the future state. So this robot follower well suitable for SLAM application . The existence of a non divergent estimation theoretic solution to the SLAM problem and to elucidate upon the general structure of SLAM navigation algorithms. These feature base SLAM contributions are maintain knowledge of the relative relationships between landmark location estimates based feature values and which in turn underpin the exhibited convergence properties.